\pdfoutput=1

\documentclass[11pt]{article}

\usepackage{emnlp2021}

\usepackage{times}
\usepackage{latexsym}

\usepackage[T1]{fontenc}

\usepackage[utf8]{inputenc}

\usepackage{microtype}

\usepackage{microtype}
\usepackage{graphicx}
\usepackage{mathrsfs}
\usepackage{dutchcal}
\usepackage{amsfonts}
\usepackage{amsmath}
\usepackage{bm}
\usepackage{multirow}
\usepackage{algorithm}
\usepackage{algorithmicx}
\usepackage{algpseudocode}
\usepackage{booktabs}

\newcommand*{\Cline}[1]{%
    \noalign{\global\setlength{\arrayrulewidth}{0.8pt}}%
    \cline{#1}%
    \noalign{\global\setlength{\arrayrulewidth}{0.4pt}}%
}

\usepackage{color}

\DeclareMathOperator*{\bert}{BERT}
\DeclareMathOperator*{\sigmoid}{Sigmoid}
\DeclareMathOperator*{\bilstm}{BiLSTM}
\DeclareMathOperator*{\mlp}{MLP}
\DeclareMathOperator*{\argmax}{arg\,max}
\DeclareMathOperator*{\avg}{Avg}

\title{Unsupervised Conversation Disentanglement through Co-Training}

\author{
    Hui Liu \qquad
    Zhan Shi \qquad 
    Xiaodan Zhu \\
    Ingenuity Labs Research Institute \& ECE, Queen’s University, Canada \\ 
    {\tt \{hui.liu, z.shi, xiaodan.zhu\}@queensu.ca }
}

\begin{document}
\maketitle
\begin{abstract}
Conversation disentanglement aims to separate intermingled messages into detached sessions, which is a fundamental task in understanding multi-party conversations.
Existing work on conversation disentanglement relies heavily upon human-annotated datasets, which are expensive to obtain in practice.
In this work, we explore to train a conversation disentanglement model without referencing any human annotations.
Our method is built upon a deep co-training algorithm, which consists of two neural networks: a message-pair classifier and a session classifier.
The former is responsible for retrieving local relations between two messages while the latter categorizes a message to a session by capturing context-aware information.
Both networks are initialized respectively with pseudo data built from an unannotated corpus.
During the deep co-training process, we use the session classifier as a reinforcement learning component to learn a session assigning policy by maximizing the local rewards given by the message-pair classifier.
For the message-pair classifier, we enrich its training data by retrieving message pairs with high confidence from the disentangled sessions predicted by the session classifier.
Experimental results on the large Movie Dialogue Dataset demonstrate that our proposed approach achieves competitive performance compared to the previous supervised methods.
Further experiments show that the predicted disentangled conversations can promote the performance on the downstream task of multi-party response selection.

\end{abstract}

\section{Introduction}

\begin{figure}[t]
  \includegraphics[width=\linewidth]{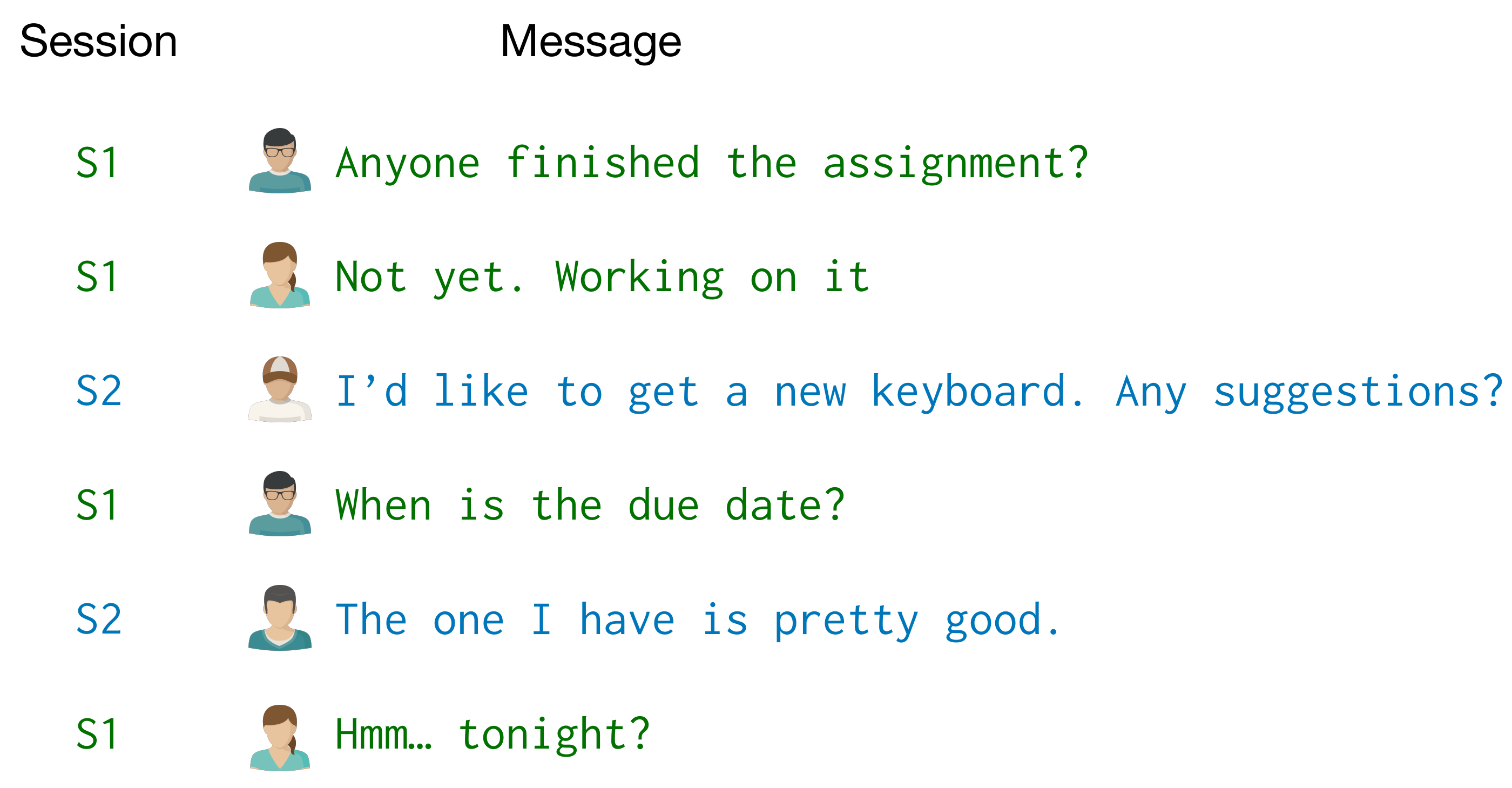}
  \caption{An example of conversation disentanglement with two sessions occurring at the same time.}
  \label{fig:dia-dis-example}
\end{figure}

With the continuing growth of Internet and social media, online group chat channels, e.g.,  Slack\footnote{https://slack.com/} and Whatsapp\footnote{https://www.whatsapp.com/}, among many others, have become increasingly popular and played a significant social and economic role.
Along with the convenience of instant communication brought by these applications, the inherent property that multiple topics are often discussed in one channel hinders an efficient access to the conversational content.
In the example shown in Figure \ref{fig:dia-dis-example}, people or intelligent systems have to selectively read the messages related to the topics they are interested in from hundreds of messages in the chat channel.

With the goal of automatically grouping messages with the same topic into one session, \textit{conversation disentanglement} has proved to be a prerequisite for understanding multi-party conversations and solving the corresponding downstream tasks such as response selection \cite{elsner2008you,lowe2017training,jia2020multi,wang2020response}. 
Previous research on conversation disentanglement can be roughly divided into two categories: (1) two-step methods, and (2) end-to-end methods.
In the two-step methods \cite{elsner2011disentangling,elsner2008you,jiang2018learning}, a model first retrieves the ``local'' relations between two messages by utilizing either feature engineering approaches or deep learning methods, and then a clustering algorithm is employed to divide an entire conversation into separate sessions based on the message pair relations.
In contrast, end-to-end methods \cite{tan2019context,yu2020online} capture the ``global'' information contained in the context of detached sessions and calculate the matching degree between a session and a message in an end-to-end manner. 
Though end-to-end methods have been proved to be more flexible and can achieve better performance \cite{ijcai2020-535}, these two types of methods are interconnected and complementary since a global optimal clustering solution on the local relations will produce the optimal disentanglement scheme \cite{NIPS2004_16808292}.

Although the previous research efforts have achieved an impressive progress on conversation disentanglement, they all highly rely on human-annotated corpora, which are expensive and scarce to obtain in practice \cite{kummerfeld2019large}.
The heavy dependence on human annotations limits the extensions of related study on conversation disentanglement as well as the applications on downstream tasks, given a wide variety of occasions where multi-party conversations can happen.
In this work, we explore the possibility to train an end-to-end conversation disentanglement model without referencing any human annotations and propose a completely unsupervised disentanglement model.

\begin{figure}[t]
  \includegraphics[width=\linewidth]{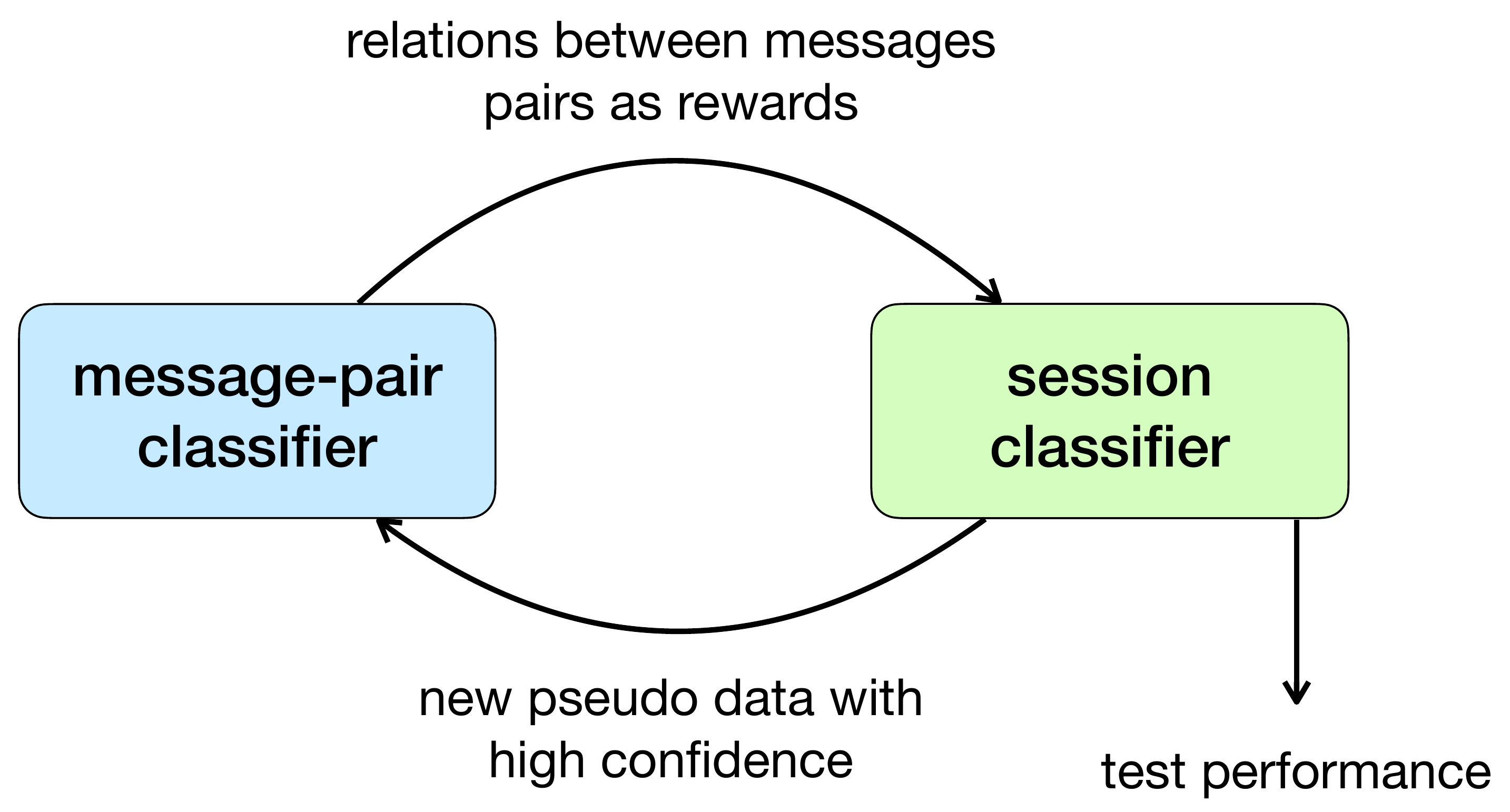}
  \caption{Illustration of our proposed co-training framework.}
  \label{fig:co-training-example}
\end{figure}

Our method builds upon the co-training approach \cite{blum1998combining,nigam2000analyzing} but extends it to a deep learning framework.
By viewing the disentanglement task from the local perspective and the global perspective, our method consists of a message-pair classifier and a session classifier. 
The message-pair classifier aims to retrieve the message pair relations, which is of a similar purpose as the model used in a two-step method that retrieves the local relations between two messages.
The session classifier is a global context-aware model that can directly categorize a message into a session in an end-to-end fashion.
The two classifiers view the task of conversation disentanglement from the perspectives of a local two-step method and an global end-to-end model, which will be separately initialized with pseudo data built from the unannotated corpus and updated with each other during co-training.
More concretely, during the co-training procedure, we adopt reinforcement learning to learn a session assigning policy for the session classifier by maximizing the accumulated rewards between a message and a session which are given by the message-pair classifier.
After updating the parameters of the session classifier, a new set of data with high confidence will be retrieved from the predicted disentanglement results of the session classifier and used for updating the message-pair classifier.
As shown in Figure \ref{fig:co-training-example}, the above process is iteratively performed by updating one classifier with the other until the performance of session classifier stops increasing.

We conduct experiments on the large public Movie Dialogue Dataset \cite{ijcai2020-535}.
Experimental results demonstrate that our proposed method outperforms strong baselines based on BERT \cite{devlin-etal-2019-bert} in two-step settings, and achieves competitive results compared to those of the state-of-the-art supervised end-to-end methods.
Moreover, we apply the disentangled conversations predicted by our method to the downstream task of multi-party response selection and get significant improvements compared to a baseline system.\footnote{Code will be publicly available at \url{https://github.com/LayneIns/Unsupervised_dialo_disentanglement}}
In summary, our main contributions are three-fold:
\begin{itemize}
\itemsep0em 
    \item To the best of our knowledge, this is the first work to investigate unsupervised conversation disentanglement with deep neural models.
    \item We propose a novel approach based on co-training which can perform unsupervised conversation disentanglement in an end-to-end fashion.
    \item We show that our method can achieve performance competitive with supervised methods on the large public Movie Dialogue Dataset. Further experiments show that our method can be easily adapted to downstream tasks and achieve significant improvements. 
\end{itemize}

\section{Related Work}

\paragraph{Conversation Disentanglement}
Conversation disentanglement has long been regarded as a fundamental task for understanding multi-party conversations \cite{elsner2008you,elsner2010disentangling} and can be combined with downstream tasks to boost their performance \cite{jia2020multi,wang2020response}.
Previous methods on conversation disentanglement are mostly performed in a supervised fashion, which can be classified as two categories: (1) two-step approaches and (2) end-to-end methods. 
The two-step methods \cite{elsner2008you,elsner2010disentangling,elsner2011disentangling,chen2017learning,jiang2018learning,kummerfeld2019large} firstly retrieve the relations between two messages, e.g., ``reply-to'' relations \cite{guo2018answering,zhu2020did}, and then adopt a clustering algorithm to construct individual sessions.
The end-to-end models \cite{tan2019context,ijcai2020-535,yu2020online}, instead, perform the disentanglement operation in an end-to-end manner, where the context information of detached sessions will be exploited to classify a message to a session.
End-to-end models tend to achieve better performance than two-step models, but both often need large annotated data to get fully trained \cite{ijcai2020-535}, which is expensive to obtain and thus encourages the demand on unsupervised algorithms.  
A few preliminary studies perform unsupervised thread detection in email systems based on two-step methods \cite{wu2005indexing,erera2008conversation,domeniconi2016novel}, but these methods use handcrafted features which cannot be extended to various datasets.
Compared with previous works, our method can conduct end-to-end conversation disentanglement in a completely unsupervised fashion, which can be easily adapted to downstream tasks and used in a wide variety of applications.

\paragraph{Dialogue Structure Learning}

One problem that may be related to conversation disentanglement is dialogue structure learning \cite{zhai2014discovering,shi2019unsupervised}.
Both are related to understanding multi-party conversation structures but they are different tasks.
Dialogue structure learning aims to discover latent dialogue topics and construct an implicit utterance dependency tree to represent a multi-party dialogue’s turn taking \cite{qiu2020structured}, while the goal of conversation disentanglement is to learn an explicit dividing scheme that separates intermingled messages into sessions.

\paragraph{Co-training}

Co-training \cite{blum1998combining,nigam2000analyzing} has been widely used as a low-resource learning algorithm in natural language processing \cite{wu2018reinforced,ijcai2018-556}, which assumes that the data has two complementary views and utilizes two models to iteratively provide pseudo training signals to each other.
Our method consists of a message-pair classifier and a session classifier, which respectively view the unannotated dataset from the perspective of the local relations between two messages and that of the context-aware relations between a session and a message.
To the best of our knowledge, this is the first work that utilizes co-training in the research of conversation disentanglement. We will extend the co-training idea to the deep learning paradigm to construct novel models for disentanglement.

\section{Formulation and Notations}

Given a conversation $\mathcal{C} = [m_1, m_2, \cdots, m_N]$ where $m_i$ is a message with speaker $\mathcal{S}_i$, it contains $K$ sessions $\{T^k\}_{k=1}^K$ where $T^k=[m^k_1, \cdots, m^k_{\vert T^k \vert}]$ and $K$ are unknown to the model.
Our goal is to learn a dividing scheme that indicates which session a message $m_i$ belongs to.
We solve this task in an end-to-end fashion where we formulate unsupervised conversation disentanglement as an unsupervised sequence labeling task.
For a given message $m_i$, there exists a session set $\mathbb{T}=\{T^1, \cdots, T^{z(i)}\}$ where $z(i)$ indicates the number of detached sessions when $m_i$ is being processed.
The model needs to decide if $m_i$ belongs to any session in $\mathbb{T}$.
If $m_i \in T^k$, then $m_i$ will be appended to $T^k$; otherwise a new session $T^{z(i)+1}$ will be built and initialized by $m_i$, and the new session will be added to $\mathbb{T}$.

\section{Method}

In this section, we describe our co-training based framework in detail, which contains following components:
\begin{enumerate}
\itemsep0em 
    \item A message-pair classifier which can retrieve the relations between two messages. The relation scores will be used as rewards for updating the session classifier during co-training.
    \item A session classifier which can perform end-to-end conversation disentanglement by retrieving the relations between a message and a session. The predicted results will be used to build new pseudo data to train the message-pair classifier during co-training.
    \item A co-training algorithm involving the message-pair classifier and the session classifier. Two classifiers will help to update each other until the performance of the session classifier stops growing.
\end{enumerate}

We will introduce the details of the three components in following sections.

\subsection{Message-pair Classifier}

The message-pair classifier is a binary classifier which we denote as $\mathcal{F}_m$ in the remainder of this paper.
Due to the lack of annotated data in unsupervised settings, the goal of $\mathcal{F}_m$ is to predict if two messages are in the same session; i.e., whether they talk about the same topic, which is different from most previous work that predicts the ``reply-to'' relation.
In our experiment, we adopt a pretrained BERT \cite{devlin-etal-2019-bert} in the base version as our message encoder.

\subsubsection{Model}

Given two messages $m_i$ and $m_j$, we separately obtain the sentence embeddings of the two messages:
\begin{gather}
    \label{eq:func1}
    v_{m_i} = \bert(m_i) \\
    v_{m_j} = \bert(m_j)
\end{gather}
The probability of $m_i$ and $m_j$ belonging to the same session is computed with the dot product between $v_{m_i}$ and $v_{m_j}$:
\begin{gather}
    \label{eq:func3}
    p_m = \sigmoid(v_{m_i} \cdot  v_{m_j})
\end{gather}
We abbreviate Eq. \ref{eq:func1}--\ref{eq:func3} as:
\begin{gather}
    p_m(m_i, m_j)=\mathcal{F}_m(m_i, m_j)
\end{gather}
$\mathcal{F}_m$ is trained to minimize the cross-entropy loss.
The predicted probabilities between message pairs will be used as rewards during the co-training process to update the session classifier.

\subsubsection{Initialization}
\label{sec:two-step-init}

One important step in standard co-training algorithm is to initialize the classifiers with a small amount of annotated data.
Since our dataset is completely unlabeled, we create a pseudo dataset to initialize the message-pair classifier. 
The assumption we use in our experiments is that one speaker mostly participates in only one session\footnote{We verify this assumption in two natural multi-party conversation datasets: Reddit dataset \cite{tan2019context} and Ubuntu IRC dataset \cite{kummerfeld2019large}. Statistics show that only 6\% of speakers will join multiple sessions on the Reddit dataset and 20\% on the IRC dataset.}.

To construct the pseudo data $\mathcal{D}_m$, we use the message pairs from the same speaker in one conversation as the positive cases, while randomly sampling messages from different conversations as the negative pairs.
In this way we obtain a retrieved dataset $\mathcal{D}_m^{ret}$ containing 937K positive cases and 2,184K negative cases.
However, we observe that the positive cases constructed from the above process are very noisy because: (1) there are still some speakers who will appear in multiple sessions, and (2) even message pairs from the same speaker in the same session can be very semantically different since they are not contiguous messages.
These noisy training cases will result in low confidences for the predicted probabilities of $\mathcal{F}_m$, which will be used later in co-training.
Thus we randomly select some messages from the unlabeled dataset and use a pretrained DialoGPT \cite{zhang2020dialogpt} to generate direct responses in order to form new positive cases, which we denote as $\mathcal{D}_m^{gen}$ .
In this way, we finally obtain the pseudo data $\mathcal{D}_m = \mathcal{D}_m^{ret} \cup \mathcal{D}_m^{gen}$, which contains 1,212K positive cases and 2,184K negative cases, to initialize $\mathcal{F}_m$.

\subsubsection{Two-step Disentanglement}



After being trained on the pseudo data $\mathcal{D}_m$, the message-pair classifier $\mathcal{F}_m$ can be exploited as for two-step conversation disentanglement.
Given an unlabeled conversation as $\mathcal{C} = [m_1, m_2, \cdots, m_N]$, we first use $\mathcal{F}_m$ to predict the probability between each message pair in $\mathcal{C}$.
Then we perform the greedy search algorithm widely used in the previous works \cite{elsner2008you} to segment $\mathcal{C}$ into detached sessions.

\subsection{Session Classifier}

The session classifier, denoted as $\mathcal{F}_t$, aims to calculate the relations between a session and a message that indicates if the message belongs to the session or not. 
Given the current context of a session as $T = [m_1, \cdots, m_{\vert T\vert}]$ and a message $m$, the goal of $\mathcal{F}_t$ is to decide if $m$ can be appended to $T$ or not.

\subsubsection{Model}

For each message $m_j \in T$, we obtain its sentence embedding $v_{m_j}$ by a Bidirectional LSTM network \cite{hochreiter1997long} and a multilayer perceptron (MLP):
\begin{gather}
    \label{eq:func5}
    \overrightarrow{v}_{m_j}, \overleftarrow{v}_{m_j} = \bilstm(m_j)\\
    \label{eq:func6}
    v_{m_j} = \mlp([\overrightarrow{v}_{m_j}, \overleftarrow{v}_{m_j}])
\end{gather}
After obtaining sentence embeddings of all the messages in $T$ as $[v_{m_1}, \cdots v_{m_{\vert T\vert}}]$, we adopt a self attention mechanism \cite{yang2016hierarchical} to calculate the session embedding $v_{T}$ by aggregating the information from different messages.
Specifically,
\begin{gather}
    u_{m_j} = \tanh(w \cdot v_{m_j} + b)\\
    \alpha_{m_j} = \frac{\exp(u_{m_j})}{\sum_j \exp(u_{m_j})}\\
    v_{T} = \sum_j \alpha_{m_j} v_{m_j}
\end{gather}
where $w$ and $b$ are trainable parameters.

For the message $m$, we use the same Bidirectional LSTM network and MLP as in Equation \ref{eq:func5} and \ref{eq:func6} to obtain its sentence embedding $v_{m}$.
Then the probability of $m$ belonging to $T$ is calculated with the dot product between $v_{m}$ and $v_{T}$:
\begin{gather}
    p_t = \sigmoid(v_{T} \cdot  v_{m})
\end{gather}
We abbreviate the above process as:
\begin{gather}
    p_t(T, m) = \mathcal{F}_t(v_{T}, v_{m})
\end{gather}
$\mathcal{F}_t$ is trained to minimize the cross-entropy loss.

\subsubsection{Initialization}

Similar to the message-pair classifier, we build a pseudo dataset $\mathcal{D}_t$ to initialize the session classifier $\mathcal{F}_t$ to make it be able to decide if a message is semantically consistent with a sequence of messages. 
We construct $\mathcal{D}_t$ based on the same assumption that one speaker is involved in just one session for most of the time.

Given a conversation $\mathcal{C} = [m_1, m_2, \cdots, m_N]$ from the unlabeled corpus, we retrieve the messages from a speaker $\mathcal{S}$ as $\mathcal{C}_{\mathcal{S}} = [m_{\mathcal{S}_1}, m_{\mathcal{S}_2}, \cdots ]$ where $\mathcal{C}_{\mathcal{S}} \subset \mathcal{C}$.
Based on the assumption, the messages in $\mathcal{C}_{\mathcal{S}}$ are in the same session, so the message $m_{\mathcal{S}_i} \in \mathcal{C}_{\mathcal{S}}$ where $i\neq 1$ and its preceding context can be regarded as the positive input of $\mathcal{F}_t$.
Consider a positive case with $m_{\mathcal{S}_2}$ as the message, the message $m$ and session $T$ are defined as follows:
\begin{gather}
    m := m_{\mathcal{S}_2} \\
    T := [m_1, \cdots, m_{\mathcal{S}_2 -1}]
\end{gather}
The reason is that $m_{\mathcal{S}_1} \in [m_1, \cdots, m_{\mathcal{S}_2 -1}]$, so $[m_1, \cdots, m_{\mathcal{S}_2 -1}]$ and $m_{\mathcal{S}_2}$ should be semantically consistent according to the assumption.

For the negative instances of $\mathcal{D}_t$, we randomly sample a conversation as $T$ from the corpus, and a message from another conversation as $m$. As such we obtain a pseudo dataset $\mathcal{D}_t$ consisting of 460K positive instances and 1,158K negative cases.

\subsubsection{End-to-end Disentanglement}

\begin{algorithm}[t]
  \caption{An end-to-end method for conversation disentanglement with the session classifier}
  \label{alg:end-to-end}
  \textbf{Input}: An unlabeled conversation $\mathcal{C}$, the initialized session classifier $\mathcal{F}_t$\\
  \textbf{Output}: A set of sessions $\mathbb{T}$
  \begin{algorithmic}[1]
    \State Let $\mathbb{T} = \emptyset$
    \For {$m_i \in \mathcal{C}$}
        \If {$\mathbb{T}$ is empty}
            \State Let $T^{z(i)+1}=\{m_i\}$
            \State $\mathbb{T} = \mathbb{T} \cup \{T^{z(i)+1}\}$
        \Else 
            \State Let $\mathcal{C}_i = [m_1, \cdots, m_{i-1}]$ \label{alg:new-start}
            \State $p_t(\mathcal{C}_i, m_i) = \mathcal{F}_t(v_{\mathcal{C}_i}, v_{m_i})$ \label{alg:new-end}
            \If {$p_t(\mathcal{C}_i, m_i) < 0.5$}
                \State Let $T^{z(i)+1}=\{m_i\}$
                \State $\mathbb{T} = \mathbb{T} \cup \{T^{z(i)+1}\}$
            \Else
                \State Let $P_m(i) = \{\}$
                \For {$T^k \in \mathbb{T}$}
                    \State $p_t(T^k, m_i) = \mathcal{F}_t(v_{T^k}, v_{m_i})$
                    \State Add $p_t(T^k, m_i)$ to $P_m(i)$
                \EndFor
                \State $T^k_{\max} = \argmax P_m(i)$
                \State $T^k_{\max} = T^k_{\max} \cup \{m_i\}$
            \EndIf
        \EndIf
    \EndFor
  \end{algorithmic}
  \textbf{Return} $\mathbb{T}$ \\
\end{algorithm}

Note that after initialized with the pseudo data $\mathcal{D}_t$, the session classifier $\mathcal{F}_t$ can be directly applied to perform end-to-end conversation disentanglement.
Suppose message $m_i$ is being processed where $m_i \in \mathcal{C}$ and $\mathcal{C} = [m_1, m_2, \cdots, m_N]$, we first calculate the probability of $m_i$ belonging to its preceding context $\mathcal{C}_i = [m_1, \cdots, m_{i-1}]$, which we denote as $p_t(\mathcal{C}_i, m_i) = \mathcal{F}_t(v_{\mathcal{C}_i}, v_{m_i})$.
If $p_t(\mathcal{C}_i, m_i) < 0.5$, $m_i$ will be used to initialize a new session $T^{z(i)+1}$ where $z$ is a function indicating the number of disentangled sessions in $\mathcal{C}_i$; otherwise $m_i$ will be used to calculate the matching probability with each session in $\mathbb{T}$, and then be classified to the session which has the greatest matching probability.
The overall end-to-end algorithm is shown in Algorithm \ref{alg:end-to-end}.

\subsection{Co-Training}

The confidence of $\mathcal{F}_m$ and $\mathcal{F}_t$ is not high because they are initialized with noisy pseudo data.
We propose to adapt the idea of co-training to the disentanglement task, which is leveraged to iteratively update the two classifiers with the help of each other.
The session classifier will utilize the local probability provided by the the message-pair classifier with reinforcement learning, while more training data, built from the outcomes of the session classifier, will be fed to the message-pair classifier.
We will introduce the details in this subsection.

\subsubsection{Updating Session Classifier}

Since no labeled data is provided to train $\mathcal{F}_t$, we formulate the disentanglement task as a deterministic Markov Decision Process and adopt the Policy Gradient algorithm \cite{sutton1999policy} for the optimization.
For each co-training iteration, $\mathcal{F}_t$ will be initialized with the pseudo data $\mathcal{D}_t$ and then updated by reinforcement learning.

\paragraph{State}

The state $s_i$ of the $i^{th}$ disentanglement step consists of three components $(m_i, \mathcal{C}_i, \mathbb{T})$, where $m_i$ is the $i^{th}$ message of $\mathcal{C}$; $\mathcal{C}_i = [m_1, \cdots, m_{i-1}]$ is the preceding context of $m_i$; $\mathbb{T}$ is the detached session set which contains $z(i)$ sessions.

\paragraph{Action}

The action space of the $i^{th}$ disentanglement step consists of two types of actions: 
\begin{enumerate}
\itemsep-0.3em 
    \item Classifying $m_i$ to a new session, which we denote as $a_i^{new} \in \{0, 1\}$. If $a_i^{new}$ is 0, $m_i$ will be used to initialize a new session $T^{z(i)+1}$; otherwise $m_i$ will be categorized into an existing session.
    \item Categorizing $m_i$ to an existing session in $\mathbb{T}$, which we denote as $a_i^t \in \{1, \cdots, z(i)\}$. 
\end{enumerate}

\paragraph{Policy network}

We parameterize the action with a policy network $\pi$ which is in a hierarchical structure.
The first layer policy $\pi^{new}(\bm{a}^{new}_i \vert s_i; \theta_1)$ is to decide if a message $m_i$ belongs to $\mathcal{C}_i$, and the first layer action $a_i^{new} \in \{0, 1\}$ will be sampled:
\begin{gather}
    \pi^{new}(\bm{a}^{new}_i \vert s_i; \theta_1) = p_t(\mathcal{C}_i, m_i) \\
    a_i^{new} \sim \pi^{new}(\bm{a}^{new}_i \vert s_i; \theta_1)
\end{gather}
If $a_i^{new}$ is 1, which means $m_i$ belongs to a session in $\mathbb{T}$, the second layer policy $\pi^t(\bm{a}^t_i \vert s_i; \theta_2)$ will decide which of existing sessions that $m_i$ should be categorized to:
\begin{gather}
    \pi^t(\bm{a}^t_i \vert s_i; \theta_2) = \{p_t(T^k, m_i) \vert T^k \in \mathbb{T}\} \\
    a_i^t \sim \pi^t(\bm{a}^t_i \vert s_i; \theta_2)
\end{gather}
where $\theta_1$ and $\theta_2$ are both parameters.

\paragraph{Reward}

The rewards are provided by the message-pair classifier $\mathcal{F}_m$.
For $a_i^{new}=0$, we want $m_i$ to be different from all the messages in $\mathcal{C}_i$.
Thus it is defined by the negative average of the probabilities between $m_i$ and all the messages in $\mathcal{C}_i$.
However, for $a_i^{new}=1$ and $a_i^t = k$, we want $m_i$ to be similar to all the messages in $T^k$, and thus the reward is defined as the average of the probabilities between $m_i$ and all the messages in $T^k$:
\begin{gather}
r_{m_i} = 
\begin{cases}
-\avg(\{\mathcal{F}_m(m_i, m_j)\vert m_j \in \mathcal{C}_i\}), \\
\quad\quad\quad\quad\quad\quad\quad\quad\quad\quad\quad\quad a_i^{new}=0 \\
\avg(\{\mathcal{F}_m(m_i, m_j)\vert m_j \in T^k\}), a_i^t=k
\end{cases}
\end{gather}

An issue associated with $r_{m_i}$ is that its confidence might be low because $\mathcal{F}_m$ is trained on noisy pseudo data.
We hence design another speaker reward $r_{\mathcal{S}_i}$ based on our assumptions.
For a message $m_i$ initializing a new session $T^{z(i)+1}$, its speaker $\mathcal{S}_i$ should not appear in $\mathcal{C}_i$; while for a message $m_i$ categorized to an existing session $T^k$, it should receive a positive reward if its speaker $\mathcal{S}_i$ appears in $T^k$:
\begin{gather}
r_{\mathcal{S}_i} = 
\begin{cases}
-1, & a_i^{new}=0 \text{~and~} \mathcal{S}_i \in \mathcal{C}_i \\
1,  & a_i^t=k \text{~and~} \mathcal{S}_i \in T^k \\
0,  & \text{~otherwise~}
\end{cases}
\end{gather}
The final reward $r_i$ for an action is calculated as:
\begin{gather}
    r_i = \gamma r_{m_i} + (1-\gamma) r_{\mathcal{S}_i}
\end{gather}

where $\gamma$ is a parameter ranged in $[0, 1]$ that balances $r_{m_i}$ and $r_{\mathcal{S}_i}$, which we set to 0.6 in experiments.
The policy network parameters $\theta_1$ and $\theta_2$ are learned by optimizing:
\begin{gather}
    J(\theta_1, \theta_2) = E_{(\pi^{new}, \pi^t)}[\sum_{i=1}^N r_i]
\end{gather}

\begin{table*}[t]
\centering
\renewcommand{\arraystretch}{1.2}
\begin{tabular}{c|c|c|c|c|c|c}  
\toprule
     Method Type                & Method        & NMI       & 1-1   & $\text{Loc}_3$   & Shen-F     & MSE$\downarrow$\\
    \cline{1-7}
    \multirow{3}{*}{Two-step}   & Vanilla BERT  & 0.0       & 40.20  & 49.26 & 50.50 & 1.9710 \\
    \cline{2-7}
                                & BERT + $\mathcal{D}_m^{ret}$    & 10.59 & 45.62 & 53.4 & 53.47 & 1.4750 \\
    \cline{2-7}
                                & BERT + $\mathcal{D}_m$    & 11.13 & 45.74 & 53.69 & 53.64 & 1.4617 \\
    \cline{2-7}
                                & BERT + $\mathcal{D}_m$ + CT    & 11.32 & 45.89 & 53.80 & 53.75 & 1.4602 \\
    \Cline{2-7}
                                & *BERT & 11.52   & 45.99 & 54.04 & 53.87 & 1.4636 \\
    \midrule
    \multirow{3}{*}{End-to-end} & Session Classifier + $\mathcal{D}_t$     & 24.96 & 54.26 & 60.66 & 59.16 & 0.8059 \\
    \cline{2-7}
                                & Session Classifier + $\mathcal{D}_t$ + CT     &   \textbf{29.71}   & \textbf{56.38} & \textbf{62.46}   & \textbf{60.44} & \textbf{0.6871} \\
    \Cline{2-7}
                                & *\citet{ijcai2020-535} & 35.30  & 57.31 & 63.27 & 64.37 & 0.5299 \\                        
\bottomrule
\end{tabular}
\caption{The results of conversation disentanglement.
        * means the method is performed in a supervised manner.
        ``CT'' represents co-training.
        ``Vanilla BERT'' represents BERT without finetuning.
        Note that $\mathcal{D}_m^{ret} \cup \mathcal{D}_m^{gen} = \mathcal{D}_m$, where $\mathcal{D}_m^{ret}$ is the pairs retrieved from the unannotated corpus and $\mathcal{D}_m^{gen}$ is generated by DialoGPT.
        $\downarrow$ means the lower the better.}
\label{tab:dis-res}
\end{table*}

\subsubsection{Updating Message-pair Classifier}

As mentioned in Section \ref{sec:two-step-init}, the pseudo data $\mathcal{D}_m$ for initializing the message-pair classifier $\mathcal{F}_m$ is noisy.
Thus we enrich $\mathcal{D}_m$ with new training instances $\mathcal{D}_m^{new}$ retrieved from the predicted disentanglement results of $\mathcal{F}_t$.

Given a conversation $\mathcal{C}$, $\mathcal{F}_t$ can predict the disentangled sessions as $\mathbb{T}=\{T^k\}_{k=1}^K$.
Given session $T^k=[m^k_1, \cdots, m^k_{\vert T^k \vert}]$  as an example, for a message $m_i^k$, we retrieve its preceding $M$ messages in $T^k$ and form $M$ pairs $\{(m^k_{i-M}, m^k_i), (m^k_{i-(M-1)}, m^k_i), \cdots, (m^k_{i-1}, m^k_i)\}$ as the new positive pseudo message pairs.
In order to raise the confidence of the newly added data, we filter out those pairs where the two messages have less than 2 overlapped tokens after removing stopwords.
For each co-training iteration, $\mathcal{F}_m$ is retrained on the data $\mathcal{D}_m \cup \mathcal{D}_m^{new}$.

\section{Experiments}

\subsection{Experimental Setup}

\subsubsection{Dataset}

A large corpus is often required for end-to-end conversation disentanglement.
In this work, we conduct experiments on the publicly available Movie Dialogue Dataset \cite{ijcai2020-535} which is built from online movie scripts.
It contains 29,669/2,036/2,010 instances for train/dev/test split with a total of 827,193 messages, where the session number in one instance can be 2, 3 or 4.
Since we make explorations in unsupervised settings, no labels are used in our training. 

\subsubsection{Implementation Details}
We adopt BERT \cite{devlin-etal-2019-bert} (the uncased base version) as the message-pair classifier.
For the session classifier, we set the hidden dimension to be 300, and the word embeddings are initialized with 300-d GloVe vectors \cite{pennington2014glove}.
For training, we use Adam \cite{DBLP:journals/corr/KingmaB14} for optimization; the learning rate is set to be 1e-5 for the message-pair classifier, 1e-4 for initializing the session classifier, and 1e-5 for updating the session classifier with reinforcement learning.
We iterate for 3 turns for co-training when the best performance is achieved on the development set.

\subsubsection{Evaluation Metrics}

Four clustering metrics widely used in the previous work \cite{elsner2008you,kummerfeld2019large,tan2019context} are adopted: Normalized mutual information (NMI), One-to-One Overlap (1-1), $\text{Loc}_3$ and Shen F score (Shen-F).
More explanations about the metrics can be found in Appendix \ref{sec:metric}.

We also report the mean squared error (MSE) between the predicted session numbers and the golden session numbers as previous work \cite{ijcai2020-535}.
This metric can measure whether the model can disentangle a given dialogue to the correct number of sessions.

\subsection{Results}

\subsubsection{Disentanglement Performance}
Table \ref{tab:dis-res} shows the results of unsupervised conversation disentanglement of different methods.
We can observe that: (1) for two-step methods, BERT has a very poor performance without finetuning, while after finetuned on our pseudo dataset, its performance gets improved with a relatively large margin.
(2) Utilizing the pseudo pairs generated by a pretrained DialoGPT can further improve the performance of BERT based on $\mathcal{D}_m^{ret}$.
We consider this is because the messages from one speaker are usually not contiguous in a conversation, while DialoGPT can directly produce a response to a message, which is beneficial to BERT on capturing the differences of two messages.
(3) During the co-training process, the pseudo pairs retrieved from the predictions of the session classifier can help BERT to achieve a performance close to that of a supervised BERT, which demonstrates the effectiveness of our proposed co-training framework.
(4) BERT finetuned with golden message pairs just has a marginal performance advantage compared to the pseudo data $\mathcal{D}_m$.
This is caused by the weakness of two-step methods in which the clustering algorithm is a performance bottleneck \cite{ijcai2020-535}.

\begin{table}[t]
\centering
\setlength{\tabcolsep}{4pt}
\begin{tabular}{c|cccc|c}  
\toprule
    Iteration       & NMI   & 1-1   & $\text{Loc}_3$   & Shen-F    & F1\\
    \cline{1-6}
    Base            & 24.96 & 54.26 & 60.66 & 59.16     & 68.26\\
    \cline{1-6}
    1               & 29.80 & 56.24 & 62.39 & 60.38     & 68.44\\
    2               & 29.87 & 56.33 & 62.40 & 60.41     & 68.47\\
    3               & 29.71 & 56.38 & 62.46 & 60.44     & 68.48\\
\bottomrule
\end{tabular}
\caption{The performance of the session classifier and the message-pair classifier in each co-training iteration.
        Columns NMI, 1-1, $\text{Loc}_3$ and Shen-F are for session classifier and Column F1 is for the message-pair classifier.
        ``Base'' represents session classifier trained on $\mathcal{D}_t$ and message-pair classifier finetuned on $\mathcal{D}_m$.}
\label{tab:co-training}
\end{table}

In general, end-to-end methods perform much better than two-steps methods as shown in the table, which is in accordance with the conclusions in previous works under supervised settings \cite{yu2020online}.
The session classifier trained on the pseudo data $\mathcal{D}_t$ can achieve a Shen F score of 59.61, which is +5.29 improvement compared to the supervised BERT in two-step settings.
This proves that the model structure and the approach to building $\mathcal{D}_t$ are effective for the task of unsupervised conversation disentanglement.
Meanwhile, our proposed co-training framework can further improve the performance of the session classifier and achieve competitive results with the current state-of-the-art supervised method.
With further updating during the co-training process, the session classifier raises the NMI score from 24.96 to 29.72 and 1-1 from 54.26 to 56.38.
Such a performance gain proves that our co-training framework is an important component in handling unsupervised conversation disentanglement.

Moreover, as we can see in the table, two-step methods have a high MSE on the predicted session numbers, but with the pseudo data $\mathcal{D}_m$, BERT can achieve performance which is much better than that without finetuning and even comparable with that finetuned on the golden pairs.
End-to-end session classifier achieves a significant improvement on the MSE by reducing it from 1.4602 to 0.8059, while our proposed co-training framework further improves it to 0.6871, which is close to the performance of the supervised model.
It demonstrates that the co-training method can help the session classifier to better understand the semantics in the conversation and thus to more accurately disentangle the conversation into sessions.

\subsubsection{Analysis of Co-training}

In this section we analyze the iteration process of co-training.
Table \ref{tab:co-training} shows the performance of session classifier in different iterations.
We also include in the last column of Table \ref{tab:co-training} the performance of the message-pair classifier on the task of pair relation prediction.

As we can see, the model performance is improved iteratively as the iteration increases.
For the first iteration, the reward $r_{m_i}$ is received from the base message-pair classifier, of which the F1 score on relation prediction is 68.26.
After the first iteration, new pseudo pairs will be retrieved from the disentanglement results and used to improve the performance of the message-pair classifier to 68.44.
Thus better reward $r_{m_i}$ will be provided to update the session classifier.
As shown in the table, with such a co-training procedure, performance of both the session classifier and the message-pair classifier are significantly enhanced.


\begin{table}[t]
\centering
\begin{tabular}{c|cccc}  
\toprule
    Type            & Hits@1    & Hits@2    & Hits@5    & MRR \\
    \cline{1-5}
    None            & 12.98     & 23.0      & 52.81     & 31.87 \\
    Ours            & 14.50     & 24.58     & 54.81     & 33.29 \\
    Gold            & 17.91     & 29.97     & 59.62     & 37.01 \\
\bottomrule
\end{tabular}
\caption{The performance on multi-party response selection with disentangled conversations.
        The first column respective stands for no disentanglement, the disentangled conversations predicted by our method and the golden disentangled conversations.}
\label{tab:res-sel}
\end{table}

\subsubsection{Performance on Response Selection}

Conversation disentanglement is a prerequisite for understanding multi-party conversations.
In this section we apply our predicted sessions to the downstream task: multi-party response selection.

We create a response selection dataset based on the Movie Dialogue Dataset.
We adopt a LSTM-based network to encode the conversations/sessions, and use attention mechanism to aggregate the information from different sessions as in \citet{jia2020multi}.
More details of the model and implementation can be found in Appendix \ref{sec:multi-res-sel}. 

The results are shown in Table \ref{tab:res-sel}.
Note that the three experiments are performed on the model with the same number of parameters.
We can see that with the disentangled conversations predicted by our method, there is a significant performance gain comparing with the baseline model.
Though golden disentanglement can bring the best performance, the annotations are usually expensive to acquire.
With our method, a disentanglement scheme can be obtained for better understanding multi-party conversations with no annotation cost.

\section{Conclusion}

This is the first work to investigate unsupervised conversation disentanglement with deep neural models.
We propose a novel approach based on co-training which consists of a message-pair classifier and a session classifier.
By iteratively updating the two classifiers with the help of each other, the proposed model attains a performance comparable to that of the state-of-the-art supervised disentanglement methods.  
Experiments on downstream tasks proves that our method can help better understand multi-party conversations.
Our method can be easily adapted to a different assumption, and also it can be extended to other low-resourced scenarios like semi-supervised settings, which we will leave as our future work.

\section*{Acknowledgements}

We would like to thank the anonymous reviewers for their valuable comments.

\bibliography{anthology,custom}
\bibliographystyle{acl_natbib}

\clearpage
\appendix

\section{Appendix}

\subsection{Metric Explanation}
\label{sec:metric}

We use four metrics in our experiments: Normalized mutual information (NMI), One-to-One Overlap (1-1), $\text{Loc}_3$ and Shen F score (Shen-F).
NMI is a normalization of the Mutual Information, which is a method for evaluation of two clusters in the presence of class labels.
1-1 describes how well we can extract whole conversations intact.
$\text{Loc}_3$ counts agreements and disagreements within a context window size 3.
Shen calculates the F-score for each gold-system conversation pair, finds the max for each gold conversation, and averages weighted by the size of the gold conversation.

\subsection{Multi-party Response Selection}
\label{sec:multi-res-sel}

Given a conversation $\mathcal{C} = [m_1, \cdots, m_N]$ and a candidate message $m$, the goal of response selection is to decide if message $m$ is correct response to the conversation $\mathcal{C}$.

We obtain the disentanglement scheme of $\mathcal{C}$ as $\mathbb{T} = \{T^1, T^2, \cdots, T^K\}$, where session $T^k = [m_1. \cdots, m_{\vert T^k\vert}]$.
For each session $T^k$, we encode each message $m^k_i$ within it by a Bidirectional LSTM network and a multilayer perceptron (MLP): 
\begin{gather}
    \label{eq:func22}
    \overrightarrow{v}_{m^k_i}, \overleftarrow{v}_{m^k_i} = \bilstm(m^k_i)\\
    \label{eq:func23}
    v_{m^k_i} = \mlp([\overrightarrow{v}_{m^k_i}, \overleftarrow{v}_{m^k_i}])
\end{gather}
After obtaining sentence embeddings of all the messages in $T^k$ as $[v_{m^k_1}, \cdots v_{m^k_{\vert T^k\vert}}]$, we adopt a self attention mechanism \cite{yang2016hierarchical} to calculate the session embedding $v_{T^k}$ by aggregating the information from different messages.
Specifically,
\begin{gather}
    u_{m^k_i} = \tanh(w \cdot v_{m^k_i} + b)\\
    \alpha_{m^k_i} = \frac{\exp(u_{m^k_i})}{\sum_i \exp(u_{m^k_i})}\\
    v_{T^k} = \sum_j \alpha_{m^k_i} v_{m^k_i}
    \label{eq:func26}
\end{gather}
where $w$ and $b$ are trainable parameters.
In this way we can acquire all the session representation as $\{v_{T^1}, v_{T^2}, \cdots, v_{T^K}\}$.
Meanwhile, we obtain the candidate message representation as $v_m$ with the same LSTM and MLP in Equation \ref{eq:func22}-\ref{eq:func23}.

We follow \citet{jia2020multi} to aggregate the information from different sessions with the attention mechanism:
\begin{gather}
    s^k = v_{T^k} \cdot v_m\\
    w^k = \frac{\exp(s^k)}{\sum_k \exp(s^k)}\\
    v_{\mathcal{C}} = \sum_k w^k v_{T^k}
\end{gather}

\begin{figure}[t]
  \includegraphics[width=\linewidth]{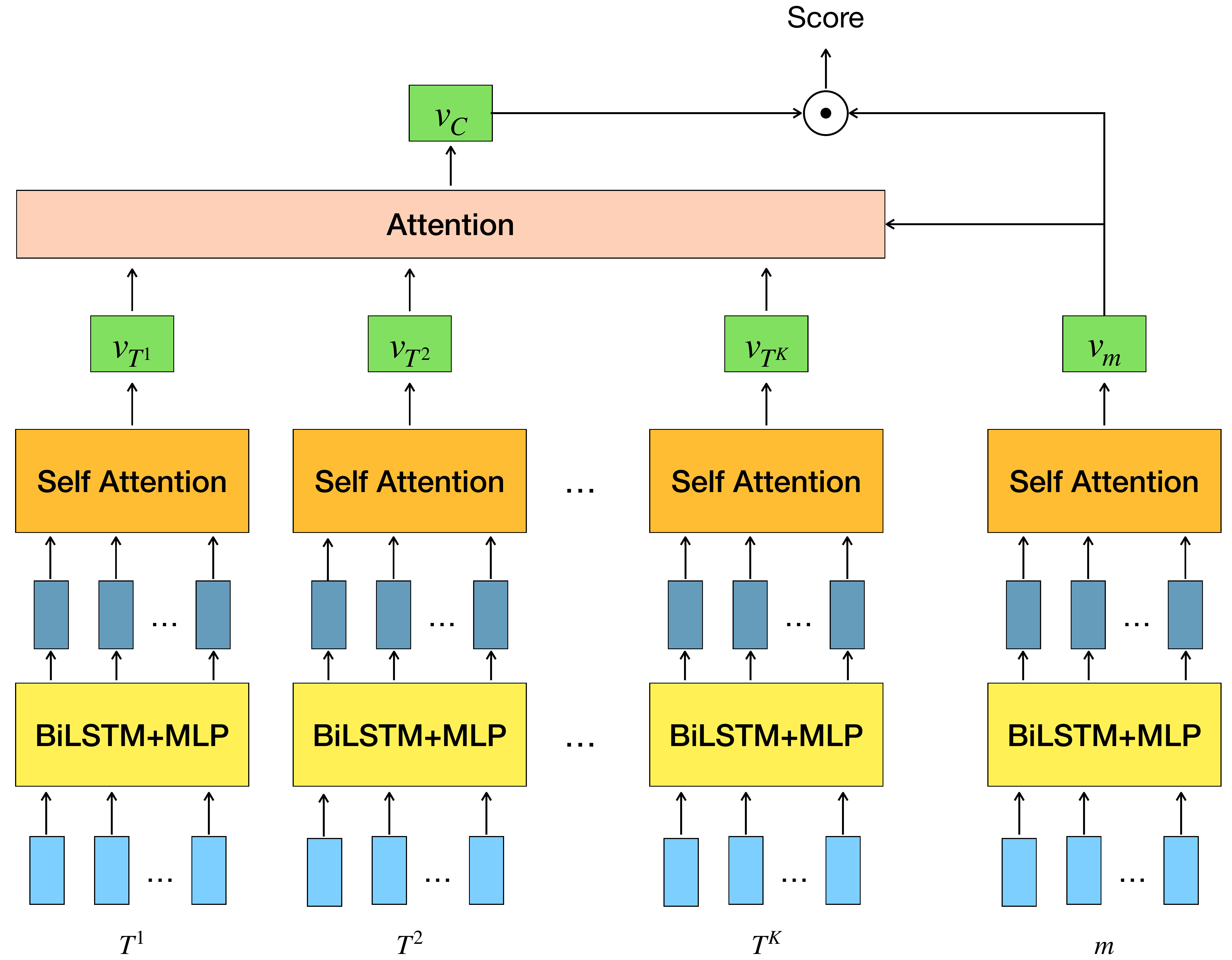}
  \caption{The model structure that incorporating disentangled sessions for the task of response selection.}
  \label{fig:res-sel-example}
\end{figure}

The final matching score between the conversation and the message is given by:
\begin{gather}
    S = v_{\mathcal{C}} \cdot v_m
\end{gather}
The overall of structure of the method incorporating the disentangled sessions is shown in Figure \ref{fig:res-sel-example}.

For the vanilla model using conversation $\mathcal{C}$ without any disentanglement, we use the LSTM, MLP and self attention as in Equation \ref{eq:func22}-\ref{eq:func26} to obtain its vector representation $v'_{\mathcal{C}}$.
And then the matching score is calculated by the dot product between $v'_{\mathcal{C}}$ and $v_m$.

The whole model is trained to minimize the cross-entropy loss of both positive instances and negative instances.

\end{document}